\documentclass[journal]{IEEEtran}

\usepackage{moreverb,url}
\usepackage{subfig}

\usepackage[colorlinks,bookmarksopen,bookmarksnumbered,citecolor=red,urlcolor=red]{hyperref}

\usepackage{graphicx}
\usepackage{subfig}
\usepackage{amsmath, amssymb, amsthm}

\usepackage{multirow} 
\usepackage[linesnumbered]{algorithm2e}
\usepackage[inline]{enumitem}

\usepackage[keeplastbox]{flushend}

\usepackage{siunitx}
\usepackage{cleveref}
\crefname{figure}{Fig.}{Fig.}
\Crefname{figure}{Fig.}{Fig.}

\usepackage{pgfplots}
\pgfplotsset{compat=newest}
\usetikzlibrary{plotmarks}
\usetikzlibrary{arrows.meta}
\usepgfplotslibrary{patchplots}
\usepackage{grffile}

\newcommand\BibTeX{{\rmfamily B\kern-.05em \textsc{i\kern-.025em b}\kern-.08em
T\kern-.1667em\lower.7ex\hbox{E}\kern-.125emX}}

\begin{document}

\title{Robot Safe Interaction System \\ for Intelligent Industrial Co-Robots}

\author{Changliu Liu and Masayoshi Tomizuka
\thanks{The work is supported by National Science Foundation Award \#1734109.}
\thanks{C. Liu, and M. Tomizuka are with the Department of Mechanical Engineering, University of California, Berkeley, CA 94720 USA (e-mail: \tt\small changliuliu, tomizuka@berkeley.edu).}
}

\maketitle

\begin{abstract}
Human-robot interactions have been recognized to be a key element of future industrial collaborative robots (co-robots). Unlike traditional robots that work in structured and deterministic environments, co-robots need to operate in highly unstructured and stochastic environments. To ensure that co-robots operate efficiently and safely in dynamic uncertain environments, this paper introduces the robot safe interaction system. In order to address the uncertainties during human-robot interactions, a unique parallel planning and control architecture is proposed, which has a long term global planner to ensure efficiency of robot behavior, and a short term local planner to ensure real time safety under uncertainties. In order for the robot to respond immediately to environmental changes, fast algorithms are used for real-time computation, i.e., the convex feasible set algorithm for the long term optimization, and the safe set algorithm for the short term optimization. Several test platforms are introduced for safe evaluation of the developed system in the early phase of deployment. The effectiveness and the efficiency of the proposed method have been verified in experiment with an industrial robot manipulator.
\end{abstract}

\begin{IEEEkeywords}
Human-Robot Interaction, Industrial Co-Robots, Motion Planning
\end{IEEEkeywords}

\section{Introduction\label{sec: intro}}

Human-robot interactions (HRI) have been recognized to be a key element of future robots in many application domains, which entail huge social and economical impacts \cite{dautenhahn2007socially, breazeal2004social}. Future robots are envisioned to function as human's counterparts, which are independent entities that make decisions for themselves; intelligent actuators that interact with the physical world; and involved observers that have rich senses and critical judgements. Most importantly, they are entitled social attributions to build relationships with humans \cite{fong2003survey}. We call these robots \textbf{co-robots}. In particular, this paper focuses on industrial co-robots.

\begin{figure}
  \begin{center}
    \includegraphics[width=8cm]{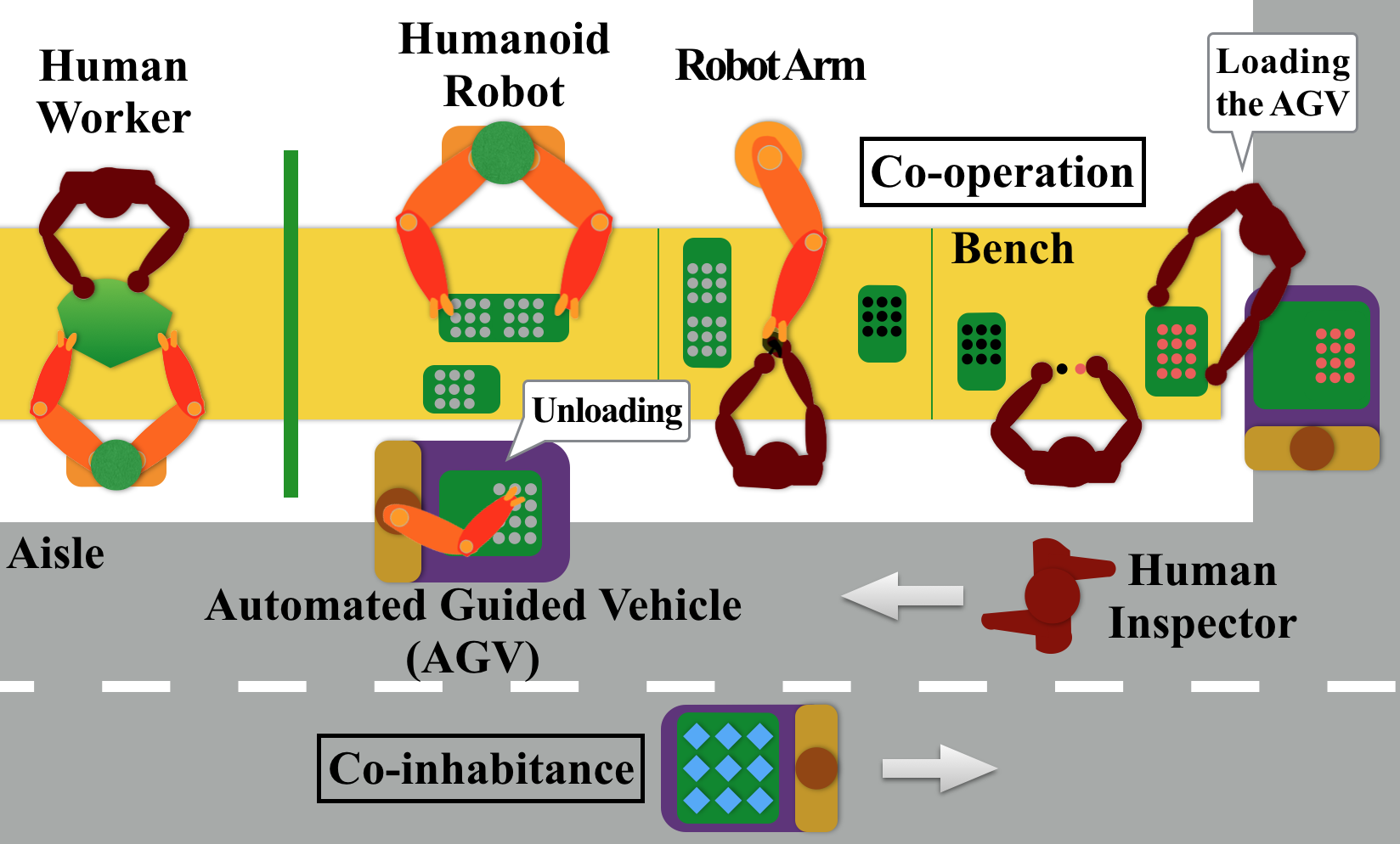}
  \end{center}
   \caption{Human-robot interactions in future production lines.\label{fig: future factory}}
\end{figure}

In modern factories, human workers and robots are two major workforces.  For safety concerns, the two are normally separated with robots confined in metal cages, which limits the productivity as well as the flexibility of production lines. In recent years, attention has been directed to remove the cages so that human workers and robots may collaborate to create a human-robot co-existing factory \cite{charalambous2013human}. Manufacturers are interested in combining human's flexibility and robot's productivity in flexible production lines \cite{koeppe2005robot}. The potential benefits of industrial co-robots are huge and extensive. They may be placed in human-robot teams in flexible production lines \cite{kruger2009cooperation} as shown in \Cref{fig: future factory}, where robot arms and human workers cooperate in handling workpieces, and automated guided vehicles (AGV) co-inhabit with human workers to facilitate factory logistics \cite{ulusoy1997genetic}. 
Automotive manufacturers Volkswagen and BMW \cite{Econ-Our-Friends-Electric} have taken the lead to introduce human-robot cooperation in final assembly lines in 2013. 

In the factories of the future, more and more human-robot interactions are anticipated to take place. Unlike traditional robots that work in structured and deterministic environments, co-robots need to operate in highly unstructured and stochastic environments. The fundamental problem is \textit{how to ensure that co-robots operate efficiently and safely in dynamic uncertain environments.}

New ISO standards has been developed concerning these new applications \cite{harper2010towards}. Several safe cooperative robots or co-robots have been released \cite{anandan2014major}, such as Green Robot from FANUC (Japan), UR5 from Universal Robots (Denmark), Baxter from Rethink Robotics (US), NextAge from Kawada (Japan), and WorkerBot from Pi4\_Robotics GmbH (Germany). 
However, many of these products focus on intrinsic safety, i.e. safety in mechanical design \cite{hirzinger2001new}, actuation \cite{zinn2004new} and low level motion control \cite{luo2011adaptive}. 
Safety during interactions with humans, which are key to intelligence (including perception, cognition and high level motion planning and control), still needs to be explored.

Technically, it is challenging to design the behavior of industrial co-robots. 
In order to make the industrial co-robots human-friendly, they should be equipped with the abilities \cite{haddadin2011towards} to:  collect environmental data and interpret such data, adapt to different tasks and different environments, and tailor itself to the human workers' needs.  
The challenges are (i) coping with complex and time-varying human motion, and (ii) assurance of real time safety without sacrificing efficiency. 

This paper introduces the robot safe interaction system (RSIS), which establishes a methodology to design the robot behavior to \textbf{safely and efficiently} interact with humans. In order to address the uncertainties during human-robot interactions, a unique parallel planning and control architecture is introduced, which has a long term global planner to ensure efficiency of robot behavior, and a short term local planner to ensure real time safety under uncertainties. The parallel planner integrates fast algorithms for real-time motion planning, e.g. the convex feasible set algorithm (CFS) \cite{liu2017sicon} for the long term planning, and the safe set algorithm (SSA) \cite{liu2014control} for the short term planning. 

An early version of the work has been discussed in \cite{liu2016algorithmic}, where the long-term planner does not consider moving obstacles (humans) in the environment. The interaction with humans is solely considered in the short-term planner. This method has low computation complexity, as the long term planning problem becomes time-invariant and can be solved offline. However, the robot may be trapped into local optima due to lack of global perspective in the long term. Nonetheless, with the introduction of CFS, long term planning can be handled efficiently in real time. Hence this paper considers interactions with humans in both planners.

This paper contributes in the following four aspects. First, it proposes an integrated framework to design robot behaviors, which includes a macroscopic multi-agent model and a microscopic behavior model. Second, it introduces a unique parallel planning architecture that integrates previously developed algorithms to handle both safety and efficiency under system uncertainty and computation limits. Third, two kinds of evaluation platforms for human-robot interactions are introduced to protect human subjects in the early phase of development. Last, the proposed RSIS has been integrated and tested on robot hardware, which validates its effectiveness. It is worth noting that safety is only guaranteed under certain assumptions of human behaviors. For a robot arm with fixed base, there is always a way to collide with it, especially intentionally. For industrial practice, RSIS should be placed on top of other intrinsic safety mechanisms mentioned earlier to ensure safety even when collision happens. Moreover, dual channel implementation of RSIS may be considered to ensure reliability. 

The remainder of the paper is organized as follows. \Cref{sec: problem} formulates the problem. \Cref{sec: rsis} proposes RSIS. \Cref{sec: ssa} discusses SSA for short term planning, while \Cref{sec: cfs} discusses CFS for long term planning. \Cref{sec: platforms} introduces the evaluation platforms. Experiment results with those platforms are shown in \Cref{sec: result}. \Cref{sec: discussion} points out future directions. \Cref{sec: conclusion} concludes the paper.

\begin{figure*}
\begin{center}
\includegraphics[width=15cm]{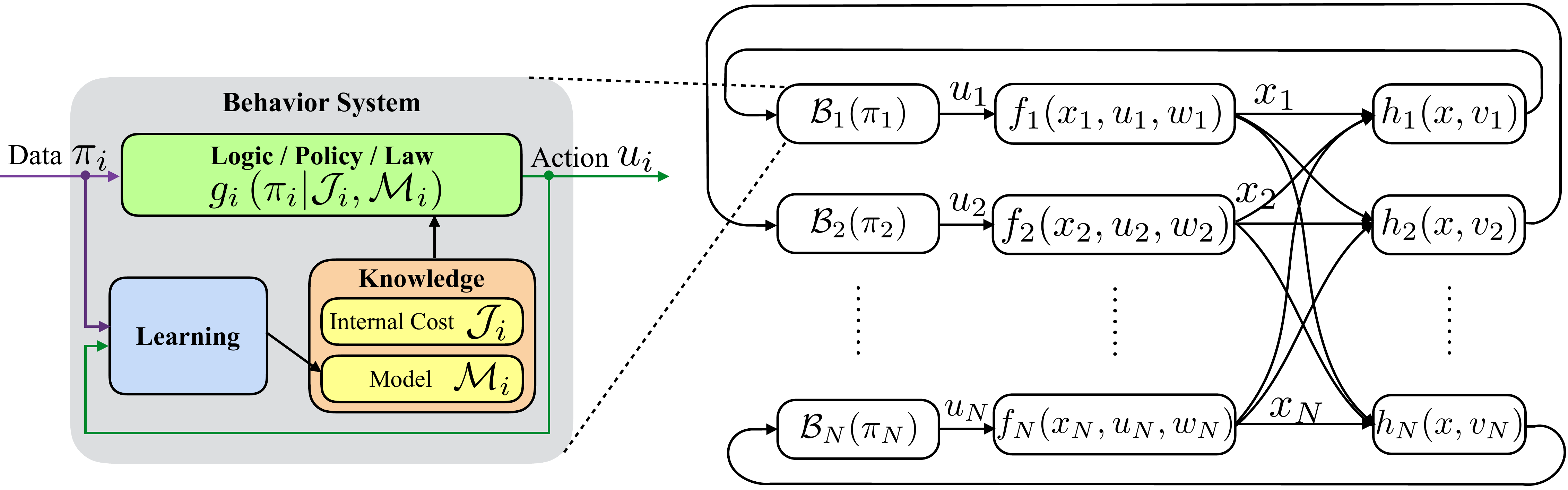}
\end{center}
\caption{The block diagram of the decomposable multi-agent system.\label{fig:Decomposable-Mutil-Agent-System}}
\end{figure*}

\section{Problem Formulation\label{sec: problem}}
Behavior is the way in which one acts or conducts oneself, especially toward others. This section introduces a multi-agent framework to model human-robot interactions. Behavior models are encoded in the multi-agent model. This paper studies the methodology of behavior design, i.e., how to realize the design goal (to ensure that co-robots operate efficiently and safely in dynamic uncertain environments) within the design scope (the inputs and outputs of the robotic system).

\subsection{Macroscopic Multi-Agent Model}
An agent is an entity that perceives and acts, whose behavior is determined by a behavior system. Human-robot interactions can be modeled in a multi-agent framework \cite{liu2014control} where robots and humans are all regarded as agents. If a group of robots are coordinated by one central decision maker, they are regarded as one agent.

Suppose there are $N$ agents in the environment and are indexed from $1$ to $N$. Denote agent $i$'s state as $x_{i}$, its control input as $u_{i}$, its data set as $\pi_{i}$ for $i=1,...,N$. The physical interpretations of the state, input and data set for different plants in different scenarios vary. For robot arm, the state can be joint position and joint velocity, and the input can be joint torque. For automated guided vehicle, the state can be vehicle position and heading, and the input can be throttle and pedal angle. When communication is considered, the state may also include the transmitted information and the input can be the action to send information. Let $x_e$ be the state of the environment. Denote the system state as $x=[x_1;\ldots;x_N; x_e]\in X$ where $X$ is the state space of the system.

Every agent $i$ has its dynamics, 
\begin{eqnarray}
\dot x_i = f_i(x_i,u_i,w_i)\label{eq: agent dynamics chap2},
\end{eqnarray}
where $w_i$ is a noise term. Agent $i$'s behavior system generates the control input $u_i$ based on the data set $\pi_{i}$,
\begin{equation}\label{eq: general_control_law chp2}
u_{i}=\mathcal{B}_i(\pi_i).
\end{equation}
where $\mathcal{B}_i$ denotes the behavior system as explained in \Cref{sec: micro behavior system} below.
Agent $i$'s data set at time $T$ contains all the observations $y_i(t)$ from the start time $t_0$ up to time $T$, i.e. $\pi_{i}\left(T\right)=\left\{ y_{i}\left(t\right)\right\} _{t\in\left[t_{0},T\right]}$ where
\begin{equation}
y_{i}=h_{i}\left(x,v_{i}\right)\label{eq: general_measurement_model chp2},
\end{equation}
and $v_{i}$ is the measurement noise. 

Based on \eqref{eq: agent dynamics chap2}, \eqref{eq: general_control_law chp2} and \eqref{eq: general_measurement_model chp2}, the closed loop dynamics become
\begin{equation}
\dot{x}  =  \mathcal{F}\left(x,w_1,\ldots,w_N,v_{1},\ldots,v_{N}\mid \mathcal{B}_1,\ldots,\mathcal{B}_N\right)\label{eq:closed_loop_N_agent},
\end{equation}
where the function $\mathcal{F}$ depends on $\mathcal{B}_i$'s. The block-diagram of the system is shown in \Cref{fig:Decomposable-Mutil-Agent-System}. 

\subsection{Microscopic Behavior System\label{sec: micro behavior system}}
To generate desired robot behavior, we need to 1) provide correct knowledge to the robot in the form of \textit{internal cost} regarding the task requirements and \textit{internal model}\footnote{Note that ``internal" means that the cost and model are specific to the designated robot.} that describes the dynamics of the environment, 2) design a correct logic to let the robot turn the knowledge into desired actions, and 3) design a learning process to update the knowledge or the logic in order to make the robot adaptable to unforeseen environments. \textit{Knowledge}, \textit{logic} and \textit{learning} are the major components of a behavior system $\mathcal{B}_i$ as shown in \Cref{fig:Decomposable-Mutil-Agent-System}. 

In the block diagram, agent $i$ obtains data $\pi_i$ from the multi-agent environment and generates its action $u_i$ according to the logic $g_i(\pi_i|\mathcal{J}_i,\mathcal{M}_i)$\footnote{The function $g_i$ is also called a control law in classic control theory or a control policy in decision theory.}, which is a mapping from information to action that depends on its knowledge, i.e., the internal cost $\mathcal{J}_i$ and the internal model $\mathcal{M}_i$. When there is no learning process, $\mathcal{B}_i(\pi_i) = g_{i}\left(\pi_{i}|\mathcal{J}_i,\mathcal{M}_i\right)$. Agent $i$'s internal model $\mathcal{M}_i$ includes the estimates of the other agents' dynamics (\ref{eq: agent dynamics chap2}) and behaviors (\ref{eq: general_control_law chp2}). 

The behavior design problem is to specify the behavior system $\mathcal{B}_i$ for the robot $i$.

\section{Robot Safe Interaction System\label{sec: rsis}}
Robot safe interaction system (RSIS) is a behavior system in order for the robot to generate safe and efficient behavior during human-robot interactions. This section first overviews the design methodology, then discusses the design considerations of the three components in the behavior system. The emphasis of this paper is the design of the logic $g_i$, which will be further elaborated in \Cref{sec: ssa,sec: cfs}.

\subsection{Overview}
There are many methods to design the robot behavior, classic control method \cite{haddadin2008collision}, model predictive control \cite{lu2015human}, adaptive control \cite{gribovskaya2011motion}, learning from demonstration \cite{tang2016teach}, reinforcement learning \cite{mnih2015human}, imitation learning \cite{kuefler2017imitating}, which ranges from nature-oriented to nurture-oriented \cite{liu2017dissertation}.
Industrial robots are safety critical. In order to make the robot behavior adaptive as well as to allow designers to have more control over the robot behavior in order to guarantee safety during human-robot interactions, a method in the framework of adaptive optimal control or adaptive model predictive control (MPC) is adopted.  
In this approach, we concern with the explicit design of the cost $\mathcal{J}_i$ and the logic $g_i$, and the design of the learning process to update $\mathcal{M}_i$, as shown in \Cref{fig:Decomposable-Mutil-Agent-System}. 

\subsection{Knowledge: The Optimization Problem}
Denote the index of the robot that we are designing for as $R$ and the index of all other agents as $H$. If $R=i$, then $x_{H}=[x_1;\ldots;x_{i-1};x_{i+1};\ldots;x_N;x_e]$. 
For robot $R$ in the multi-agent system, its internal cost is designed as,
\begin{subequations}\label{eq: chap2 design of knowledge}
\begin{align}
\mathcal{J}_R =~&  J_R(x,u_R)\label{eq: chap2 behavior design cost},\\
s.t. ~& u_R \in \Omega, x_R \in \Gamma, \dot x_R = f(x_R) + g(x_R)u_R\label{eq: chap2 behavior design self constraint},\\
& y_R = h_R(x)\label{eq: chap2 behavior design measurement constraint},\\
& x\in {X}_S\label{eq: chap2 behavior design interaction constraint},
\end{align}
\end{subequations}
where the right hand side of (\ref{eq: chap2 behavior design cost}) is the cost function for task performance. 
Equation (\ref{eq: chap2 behavior design self constraint}) is the constraint for the robot itself, i.e., constraint on the control input (such as control saturations), constraint on the state (such as joint limits for robot arms), and the dynamic constraint which is assumed to be affine in the control input. Equation (\ref{eq: chap2 behavior design measurement constraint}) is the measurement constraint, which builds the relationship between the state $x$ and the data set $\pi_R$. The noise terms are all ignored. Equation (\ref{eq: chap2 behavior design interaction constraint}) is the constraint induced by interactions in the multi-agent system, e.g. safety constraint for collision avoidance, where ${X}_S\subset X$ is a subset of the system's state space. Equation (\ref{eq: chap2 behavior design cost}) can also be called the soft constraint and (\ref{eq: chap2 behavior design self constraint}-\ref{eq: chap2 behavior design interaction constraint}) the hard constraints. Constraints (\ref{eq: chap2 behavior design self constraint}) and (\ref{eq: chap2 behavior design measurement constraint}) are determined by the physical system, while $J_R$ in (\ref{eq: chap2 behavior design cost}) and ${X}_S$ in (\ref{eq: chap2 behavior design interaction constraint}) need to be designed.

The internal model $\mathcal{M}_R$ of the robot is an estimate of others. 
However, such model is hard to obtain in the design phase for a human-robot system. They will be identified in the learning process. Moreover, since only the closed loop dynamics of others matter, we only need to identify the following function,
\begin{eqnarray}
\dot x_{H} = F_R(x_{H},x_R).\label{eq: chap2 reactive model}
\end{eqnarray}
The model $\mathcal{M}_R$ is an estimate of $F_R$, i.e., $\mathcal{M}_R = \hat {F_R}$.

\subsubsection{Example: Human-Robot Co-Operation.}

\begin{figure}[t]
\begin{center}
\includegraphics[width=4cm]{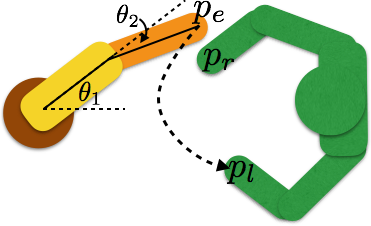}
\caption{Human-robot co-operation: One example.}
\label{fig: human robot cooperation}
\end{center}
\end{figure}

The design of the internal cost is illustrated below through an example shown in \Cref{fig: human robot cooperation}, where the human accepts a workpiece from the robot arm using his left hand. The robot arm has two links with joint positions $\theta_1$ and $\theta_2$ and the joint velocities $\dot\theta_1$ and $\dot\theta_2$. The configuration of the robot is $\theta=[\theta_{1},\theta_{2}]^{T}$. Considering the kinematics of the robot arm, the state is defined as $x_R=[\theta;\dot\theta]$ and the control input is defined to be the joint accelerations $u_{R}=\ddot{\theta}$. Then the dynamic equation satisfies,
\begin{eqnarray}
f(x_R)=\left[\begin{tabular}{cc}$0$&$I_2$\\$0$&$0$\end{tabular}\right]x_R,~g(x_R)=\left[\begin{tabular}{c}$0$\\$I_2$\end{tabular}\right]\nonumber.
\end{eqnarray}
Due to joint limits, the state constraint is $\Gamma:=\{x_R: \theta_1\in[-2\pi/3,2\pi/3],\theta_2\in[-\pi/2,\pi/2]\}$.

The area occupied by the robot arm is denoted as $\mathcal{C}(\theta)$. Denote the position of the human's left hand as $p_l$ and the position of the human's right hand as $p_r$. Then the human state is $x_{H} = [p_l;p_r;\dot p_l;\dot p_r]$, which can be predicted by (\ref{eq: chap2 reactive model}). The inverse kinematic function from the robot endpoint to the robot joint is denoted as $\mathcal{D}$. The cost function in (\ref{eq: chap2 behavior design cost}) is then designed as
\begin{equation}
J_R = \int_0^T (\|\theta(t)-\mathcal{D}(p_l(t))\|^2 + \|u_R(t)\|^2) dt,\label{eq: cost planar arm}
\end{equation}
which penalizes the distance from the robot end point to the human's left hand. The look-ahead horizon $T$ can either be chosen as a fixed number or as a decision variable that should be optimized up to the accomplishment of the task. The interactive constraint in (\ref{eq: chap2 behavior design interaction constraint}) is designed as
\begin{equation}
{X}_S(t) = \{x(t): \min_{p\in\mathcal{C}(\theta(t))} \|p_r(t) - p\| \geq d_{min}\},\label{eq: safety constraint planar arm}
\end{equation}
where $d_{min}>0$ is the minimum required distance. For simplicity, the robot only avoids the human's right hand in this formulation. This example will be re-visited in the following discussions.

\subsection{Logic: The Parallel Motion Planning\label{sec: logic}}

\begin{figure}[t]
\begin{center}
\includegraphics[width=6.5cm]{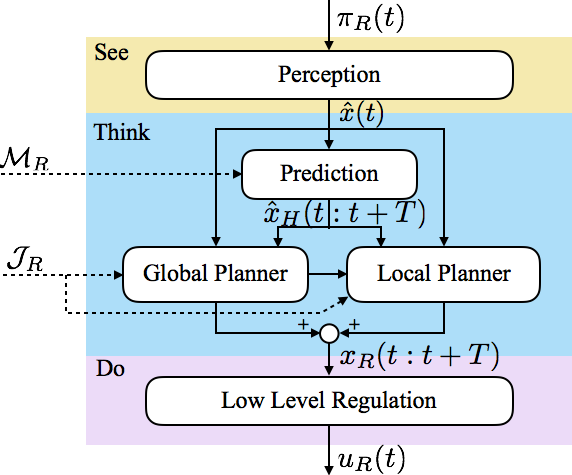}
\caption{The designed architecture of the logic in the RSIS.}
\label{fig: chap2 see think do}
\end{center}
\end{figure}

A ``see-think-do" structure shown in \Cref{fig: chap2 see think do} is adopted to tackle the complicated optimization problem (\ref{eq: chap2 design of knowledge}). The mission of the ``see" step is to process the high dimensional data $\pi_R(t)$ to obtain an estimate of the system state $\hat{x}(t)$. The ``think" step is to solve (\ref{eq: chap2 design of knowledge}) using the estimated states, and construct a realizable plan $x_R(t:t+T)$. The ``do" step is to realize the plan by generating a control input $u_R(t)$.

In the ``think" step, two important modules are \textit{prediction} and \textit{planning}. In the prediction module, robot $R$ makes predictions of other agents $\hat x_{H}(t:t+T)$ based on the internal model $\mathcal{M}_R$. The prediction can be a fixed trajectory or a function depending on robot $R$'s future state $x_R(t:t+T)$. In the planning module, the future movement $x_R(t:t+T)$ is computed given the current state $\hat x(t)$, predictions of others' motion $\hat x_{H}(t:t+T)$ as well as the task requirements and constraints encoded in $\mathcal{J}_R$. 

As it is computationally expensive to obtain the optimal solution of (\ref{eq: chap2 design of knowledge}) for all scenarios in the design phase, the optimization problem is computed in the execution phase given real time measurements. However, there are two major challenges in real time motion planning. The first challenge is the difficulty in planning a safe and efficient trajectory when there are large uncertainties in other agents' behaviors. As the uncertainty accumulates, solving the problem (\ref{eq: chap2 design of knowledge}) in the long term might make the robot's motion very conservative. The second challenge is real-time computation with limited computation power since problem (\ref{eq: chap2 design of knowledge}) is highly non-convex. We design a unique parallel planning and control architecture to address the first challenge as will be discussed below, and develop fast online optimization solvers to address the second challenge as will be discussed in the \Cref{sec: ssa,sec: cfs}.

The parallel planner consists of a global (long term) planner as well as a local (short term) planner to address the first challenge by leveraging the benefits of the two planners. The idea is to have a long term planner solving \eqref{eq: chap2 design of knowledge} while considering only rough estimation of human's future trajectory, and have a short term planner addressing uncertainties and enforcing the safety constraint \eqref{eq: chap2 behavior design interaction constraint}. The long term planning is efficiency-oriented and can be understood as deliberate thinking, while the short term planning is safety-oriented and can be understood as a reflex behavior.

\begin{figure}[t]
\begin{center}
\includegraphics[width=8.5cm]{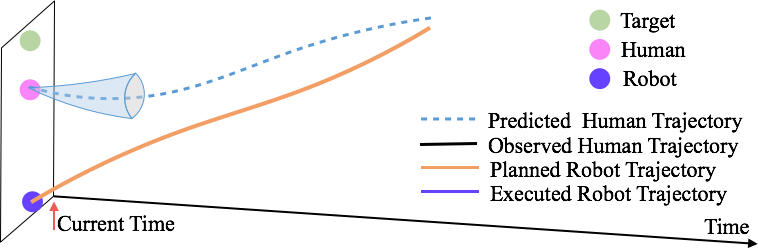}\\
\includegraphics[width=8.5cm]{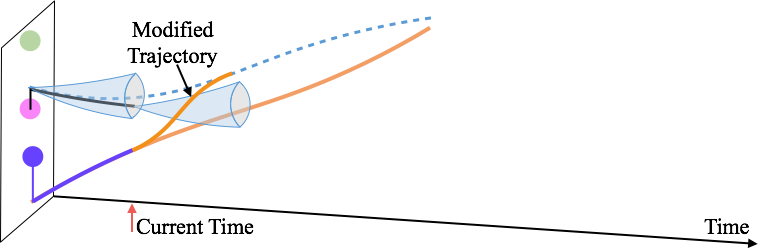}\\
\includegraphics[width=8.5cm]{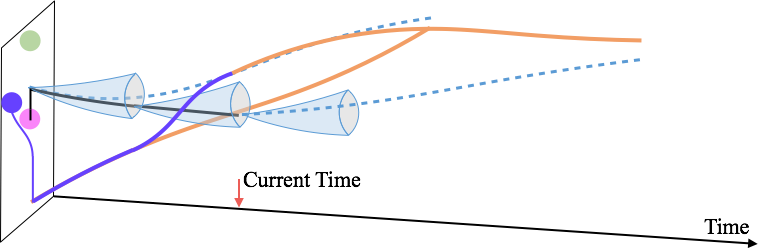}\\
\vspace{-10pt}
\subfloat[Illustration of the performance.\label{fig: parallel planning}]{
\includegraphics[width=8.5cm]{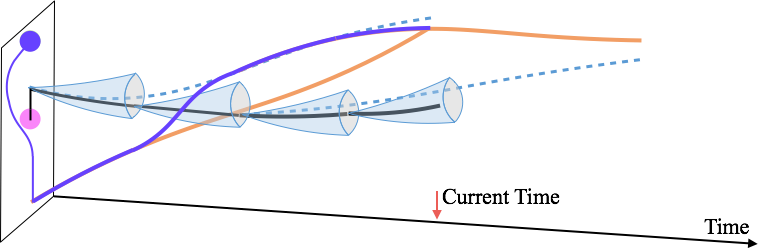}
}\\
\subfloat[Time flow of planning, execution, and computation.\label{fig: parallel time flow}]{\includegraphics[width=8.5cm]{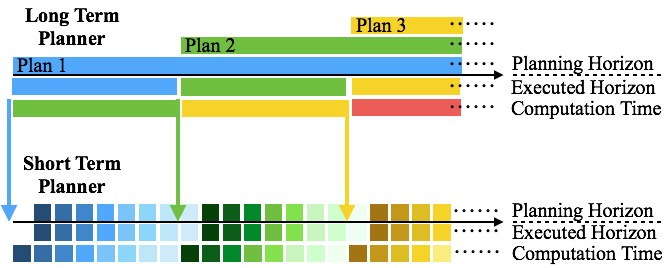}}
\caption{Parallel planning.}
\end{center}
\end{figure}

The idea is illustrated in \Cref{fig: parallel planning}. There is a closed environment. The robot (the purple dot) is required to approach the target (the green dot) while avoiding the human (the pink dot). The time axis is introduced to illustrate the spatiotemporal trajectories. For the long term planning, the robot has a rough estimation of the human's trajectory and it plans a trajectory without considering the uncertainties in the prediction. Then the trajectory is used as a reference in the short term planning. In the first time step, the robot predicts the human motion (with uncertainty estimation) and checks whether it is safe to execute the long term trajectory. As the trajectory does dot intersect with the uncertainty cone, it is executed. In the next time step, since the trajectory is no longer safe, the short term planner modifies the trajectory by detouring. Meanwhile, the long term planner comes with another long term plan and the previous plan is overwrote. The short term planner then monitors the new reference trajectory. The robot follows the trajectory and finally approaches the goal. This approach addresses the uncertainty and is non conservative. As a long term planning module is included, it avoids the local optima problem that short term planners have.

The block diagram of the parallel controller architecture is shown in \Cref{fig: chap2 see think do}. In the following discussions, we also call the long term module as the efficiency controller and the short term module as the safety controller. This approach can be regarded as a two-layer MPC approach. The computation time flow is shown in \Cref{fig: parallel time flow} together with the planning horizon and the execution horizon. Three long term plans are shown, each with one distinct color. The upper part of the time axis shows the planning horizon. The middle layer is the execution horizon. The bottom layer shows the computation time. Once computed, a long term plan is sent to the short term planner for monitoring. The mechanism in the short term planner is similar to that in the long term planner. The planning horizon, the executed horizon and the computation time for the same short term plan are shown in the same color. The sampling rate in the short term planner is much higher than that in the long term planner, since a short term plan can be computed with shorter time. In this way, the uncertainty in the short term is much smaller than it could accumulate in the long term. The planning horizon in the short term planner is not necessarily one time step, though the execution horizon is one time step. 

Coordination between the two layers is important. To avoid instability, a margin is added in the safety constraint in the long term planner so that the long term plan will not be revoked by the short term planner if the long term prediction of the human motion is correct. Theoretical analysis of the stability of the two-layer MPC method is out of the scope of this paper, and is left as a topic for future study. Nonetheless, successful implementation of the parallel architecture  depends on computation. It is important that the optimization algorithms find  feasible and safe trajectories within the sampling time. The algorithms will be discussed in \Cref{sec: ssa,sec: cfs}.

\subsection{Learning on the Model}
The internal model $\mathcal{M}_R$ needs to be obtained from the learning process. Regarding (\ref{eq: chap2 reactive model}), this paper adopts a feature-based reactive model,
\begin{eqnarray}
\dot x_H &=&  \sum_j a_{j} \varphi_{j}(x),\label{eq: chap3 reactive model adaptation}
\end{eqnarray}
where  $\varphi_{j}(x)$ are features such as distance to the robot, $a_{j}\in\mathbb{R}$ are coefficients that can be adapted online using recursive least square parameter identification (RLS-PAA) method \cite{liu2014modeling, liu2016algorithmic}. Moreover, it is assumed that $\dot x_H$ is bounded, since the maximum speed and acceleration of human are bounded.


\section{The Safety-Oriented Short Term Planning\label{sec: ssa}}
This section discusses how we use local planning to address safety. Suppose a reference input $u_R^o$ is received from the long term planner, the short term planner needs to ensure that the interaction constraint $x\in {X}_S$ will be satisfied after applying this input. Assuming $\Omega = \{u_R:|u_R|\leq \bar{u}_{R}\}$. The problem can be formulated as the following optimization,
\begin{subequations}\label{eq: chap4 optimization for local planning}
\begin{align}
u_R^* = ~& \min_{u_R} \|u_R-u_R^o\|^2_Q,\\
s.t.~& u_R\in\Omega,x_R\in \Gamma,\dot x_R = f(x_R) + h(x_R)u_R\label{eq: chap4 optimization for local planning constraint},\\
& x\in {X}_S,
\end{align}
\end{subequations}
where $\|u_R-u_R^o\|_Q=(u_R-u_R^o)^T Q(u_R-u_R^o)$ penalizes the deviation from the reference input, where $Q$ should be designed as a second order approximation of $J_R$, e.g. $Q\approx d^2 J_R/d(u_R)^2$. The constraints are the same as the constraints in (\ref{eq: chap2 design of knowledge}).  Note that modifying the control input is equivalent to modifying the trajectory as illustrated in \Cref{fig: parallel planning}. 
In this section, we call $X_S$ the safe set. Without loss of generality, it is assumed that $x\in X_S$ implies that $x_R\in\Gamma$. Otherwise, we just take the intersection of the two constraints. The safe set and the robot dynamics impose nonlinear and non-convex constraints which make the problem hard to solve. We then transform the non-convex state space constraint into convex control space constraint using the idea of invariant set. Code is available \texttt{github.com/changliuliu/SafeSetAlgorithm}.

\subsection{The Safety Principle}
Suppose the estimate of human state is $\hat x_H$ and the uncertainty range is $\Gamma_H$. According to the safe set $X_{S}$, define the state space constraint $R_S$ for the robot to be
\begin{subequations}
\begin{align}
R_S= & \{x_R: [x_R;x_H]\in X_S,\forall x_H\in \Gamma_H\}.\label{eq: robot uncertainty}
\end{align}
\end{subequations}

\textit{The safety principle} specifies that the robot control input $u_R(t)$ should be chosen such that $X_S$ is invariant, i.e. $x(t)\in X_S$ for all $t$, or equivalently,  $x_R(t)\in R_S(t)$ for $\Gamma_H(t)$ which accounts for almost all possible noises $v_1,...v_N,w_1,...,w_N$ and human behaviors $\mathcal{B}_i(\cdot),i\in H$ (those with negligible probabilities will be ignored). 

\begin{figure}[t]
\begin{center}
\includegraphics[width=5.0cm]{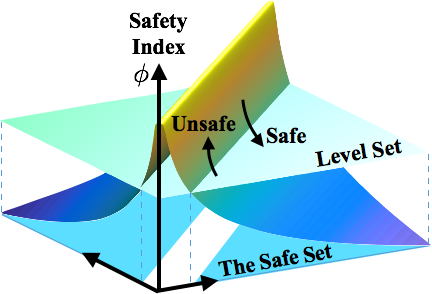}
\caption{Illustration of the safe set $X_S$ and the safety index $\phi$.\label{fig: safe set}} 
\end{center}
\end{figure}



\subsection{The Safety Index\label{sec: safety index}}

The safety principle requires the designed control input to make the safe set invariant with respect to time. In addition to constraining the motion in the safe region $R_S$, the robot should also be able to cope with any unsafe human movement. To cope with the safety issue dynamically, a safety index is introduced as shown in \Cref{fig: safe set}. The safety index $\phi:X\rightarrow\mathbb{R}$ is a function on the system state space such that
1) $\phi$ is differentiable with respect to $t$, i.e. $\dot\phi=(\partial\phi/\partial x )\dot x$ exists everywhere;
2) $\partial{\dot\phi}/\partial{u_R} \neq 0$;
3) The unsafe set $X\setminus X_S$ is not reachable given the control law $\dot\phi<0\left. when\right. \phi\geq 0$  and the initial condition $x(t_0)\in X_S$. 

The first condition is to ensure that $\phi$ is smooth. The second condition is to ensure that the robot input can always affect the safety index. The third condition provides a criteria to determine whether a control input is safe or not, e.g., all the control inputs that drive the state below the level set $0$ are safe and unsafe otherwise. The construction of such an index is discussed in \cite{liu2014control}. Applying such a control law will ensure the invariance of the safe set, hence guarantee the safety principle. To consider the constraints on control, we need to further require that $\min_{u_R\in\Omega} \dot\phi < 0$ for $\phi=0$ given that the human movement $\dot x_H$ is bounded.


\subsection{The Set of Safe Control}
To ensure safety, the robot's control must be chosen from the set of safe control
$U_S(t)=\{u_R(t):\dot\phi\leq-\eta_R\left. when\right. \phi\geq 0\}$
where $\eta_R\in \mathbb{R}^+$ is a safety margin. By the dynamic equation in (\ref{eq: chap4 optimization for local planning constraint}), the derivative of the safety index can be written as $\dot\phi=\frac{\partial \phi}{\partial x_R}hu_R+\frac{\partial \phi}{\partial x_R}f+\frac{\partial\phi}{\partial x_{H}}\dot{x}_{H}$. 
Then
\begin{equation}
U_{S}\left(t\right)=\left\{ u_{R}\left(t\right):L\left(t\right)u_{R}\left(t\right)\leq \min_{\dot x_H\in\dot\Gamma_H}S\left(t,\dot x_H\right) \right\} \label{eq:simplified safe control},
\end{equation}
where $\dot\Gamma_H$ is the set of velocity vectors that move $x_H$ to $\Gamma_H$, which can be obtained by estimating the mean squared error in the learning model \cite{liu2015safe},
\begin{subequations}
\begin{align}
L\left(t\right) =& \frac{\partial\phi}{\partial x_{R}}h\label{eq: l(t)},\\
S\left(t,\dot x_H\right) =& 
\begin{cases}
\begin{array}{c}
-\eta_{R}-\frac{\partial\phi}{\partial x_{H}}\dot{x}_{H}-\frac{\partial\phi}{\partial x_{R}}f\\
\infty\\
\end{array}
\begin{array}{c}
\phi\geq 0\\
\phi<0\\
\end{array}
\end{cases}.\label{eq: s(t)}
\end{align}
\end{subequations}

$L(t)$ is a vector at the ``safe'' direction, while $S(t,\dot x_H)$ is a scalar indicating the allowed range of safe control input, which can be broken down into three parts: a margin $-\eta_R$, a term to compensate human motion $-\frac{\partial\phi}{\partial x_{H}}\dot{x}_{H}$ and a term to compensate the inertia of the robot itself $-\frac{\partial\phi}{\partial x_{R}}f_{R}$. In the following discussion when there is no ambiguity, $S(t,\dot x_H)$ denotes the value in the case $\phi\geq 0$ only.

The difference between $R_S$ and $U_S$ is that $R_S$ is static as it is on the state space, while $U_S$ is dynamic as it concerns with the ``movements''.
Due to introduction of the safety index, the non-convex state space constraint $R_S$ is transformed to a convex control space constraint $U_S$. Since $\Omega$ is usually convex, the problem (\ref{eq: chap4 optimization for local planning}) is transformed to a convex optimization,
\begin{subequations}\label{eq: convex optimization for local planning}
\begin{align}
u_R^* = ~& \min \|u_R-u_R^o\|^2_Q,\\
s.t.~& u_R\in\Omega\cap U_S\label{eq: convex optimization for local planning constraint}.
\end{align}
\end{subequations}

The switch condition in \eqref{eq: s(t)} may result in oscillation for discrete time implementation. 
A smoothed version of the algorithm is discussed in \cite{lin2017real}.


\subsection{Example: Safe control on a planar robot arm\label{sec: ssa example}}
In this part, the design of the safety index and the computation of $U_S$ with respect to (\ref{eq: safety constraint planar arm}) will be illustrated on the planar robot arm shown in \Cref{fig: human robot cooperation}. 
The closest point to $p_r$ on the robot arm is denoted as $c\in\mathbb{R}^2$. The distance is denoted as $d = \|c-p_r\|$. The relationship among $\dot c$, $\ddot c$, $x_R = [\theta;\dot\theta]$ and $u_R = \ddot\theta$ is 
\begin{eqnarray*}
\dot c = J_c\dot\theta,~~\ddot c = J_c u_R+H_c,
\end{eqnarray*}
where $J_c$ is the Jacobian matrix at $c$ and $H_{c}=\dot{J_{c}}\dot\theta$. The constraint in (\ref{eq: safety constraint planar arm}) requires that $d\geq d_{min}$. Since the order from $c$ to $u_R$ in the Lie derivative sense is two, the safety index is designed to include the first order derivative of $d$ in order to ensure that $\partial\dot{\phi}/\partial u_R\neq 0$. The safety index is then designed as $\phi=D-d^2-k_{\phi}\dot d$.  The parameters $D>d_{min}^2$ and $k_{\phi}>0$ are chosen such that $\min_{|u_R|\leq \bar u_R} \dot\phi < 0$ for $\phi=0$ and $c$ being the robot end point. In addition to $\dot p_r$ and $\ddot p_r$ being bounded, $p_r$ is also assumed to be bounded as the human hand is not allowed to get close to the robot base. Moreover, the robot is not allowed to go to the singular point. For such $\phi$, it is easy to verify that the conditions specified above are all satisfied \cite{liu2014control}. Now we compute the constraint on the control space. Let the relative distance, velocity and acceleration vectors be $\mathbf{d}=c-p_r$,$\mathbf{v}=\dot c - \dot p_r$ and $\mathbf{a}=\ddot c - \ddot p_r$. Then $d=|\mathbf{d}|$ and
\begin{subequations}
\begin{align}
\dot\phi=&-2d\dot d-k_{\phi}\ddot d=-2\mathbf{d}^T\mathbf{v}-k_{\phi}\frac{\mathbf{d}^T\mathbf{a}+\mathbf{v}^T\mathbf{v}-\dot d^2}{d}\nonumber,\\
=&-2\mathbf{d}^T\mathbf{v}-k_{\phi}\frac{\mathbf{d}^T\left(J_c u_R+H_c-\ddot p_r\right)+\mathbf{v}^T \mathbf{v}-\frac{(\mathbf{d}^T\mathbf{v})^2}{d^2}}{d}.\nonumber
\end{align}
\end{subequations}
Hence the parameters in (\ref{eq:simplified safe control}) are
\begin{subequations}
\begin{align}
L(t) =& -k_{\phi}\frac{\mathbf{d}^T}{d}J_c,\nonumber\\
S(t) =& -\eta_R+2\mathbf{d}^T\mathbf{v}+k_{\phi}\frac{\mathbf{d}^TH_c-\mathbf{d}^T\ddot p_r+\mathbf{v}^T \mathbf{v}-\frac{(\mathbf{d}^T\mathbf{v})^2}{d^2}}{d}.\nonumber
\end{align}
\end{subequations}
Then $U_S$ can be obtained given the uncertainty on human motion $\ddot p_r$.

\section{Efficiency-Oriented Long Term Planning\label{sec: cfs}}


To avoid local optima in short term planning, long term planning is necessary. However, the optimization problem (\ref{eq: chap2 design of knowledge}) in a clustered environment is usually highly nonlinear and non-convex, which is hard to solve in real time using generic non-convex optimization solvers such as sequential quadratic programming (SQP) \cite{boggs1995sequential} and interior point method (ITP) \cite{vanderbei1999interior} that neglect the unique geometric features of the problem. CFS \cite{liu2017sicon} has been proposed to convexify the problem considering the geometric features for fast real time computation. This section further assumes that the cost function is convex with respect to the robot state and control input, and the system dynamics are linear. A method to convexify a problem with nonlinear dynamic constraints is discussed in \cite{liu2017real}. CFS code is available \texttt{github.com/changliuliu/CFS}. 

\subsection{Long Term Planning Problem\label{sec: traj smoothing problem formulation}}
Discretize the robot trajectory into $h$ points and define $x_R^q:=x_R(qt_s+t_0)$ where $t_0$ is the current time, $t_s$ is the sampling time.
The discrete trajectory is denoted $\mathbf{x}_R: =[x_R^0;x_R^1;\ldots;x_R^h]$. Similarly, the trajectory of the control input is $\mathbf{u}_R : = [u_R^0;u_R^1;\ldots;u_R^{h-1}]$. The predicted human trajectory is denoted $\hat{\mathbf{x}}_H : =[\hat x_H^0;\hat x_H^1;\ldots;\hat x_H^h]$. It is assumed that the system dynamics are linear and observable. Hence $\mathbf{u}_R$ can be computed from $\mathbf{x}_R$, i.e. $\mathbf{u}_R = \mathcal{L}(\mathbf{x}_R)$ for some linear mapping $\mathcal{L}$.   
Rewriting (\ref{eq: chap2 design of knowledge}) in the discrete time as
\begin{equation}\label{eq: benchmark smoothing chp 5}
\min_{\mathbf{x}_R\in \Gamma^e}~J_R^d(\mathbf{x}_R)
\end{equation}
where $J_R^d(\mathbf{x}_R)$ is the discretized cost function, and $\Gamma^e : = \{\mathbf{x}_R: \mathcal{L}(\mathbf{x}_R)\in\oplus_h\Omega, x_R^q\in \Gamma,[x_R^q;\hat{x}_H^q]\in {X}_S\}$. The symbol $\oplus$ is for direct sum, which is taken since $\mathbf{u}_R=\mathcal{L}(\mathbf{x}_R)$ has $h$ components and each  belongs to $\Omega$. Since $J_R$ is convex and $\mathbf{u}_R$ depends linearly on $\mathbf{x}_R$, $J_R^d$ is also convex. The non-convexity mainly comes from the constraint $\Gamma^e$. 



\subsection{Convex Feasible Set Algorithm\label{sec: application}}
To make the computation more efficient, we transforms \eqref{eq: benchmark smoothing chp 5} into a sequence of convex optimizations by obtaining a sequence of convex feasible sets inside the non-convex domain $\Gamma^e$. The general method in constructing convex feasible set is discussed in \cite{liu2017sicon}. In this paper, we assume that $\Omega$ and $\Gamma$ are  convex. The only constraint that needs to be convexified is the interactive constraint $[x_R^q;\hat{x}_H^q]\in {X}_S$. 
For each time step $q$, the infeasible set in the robot's state space is $\mathcal{O}_q:=\{x_R^q:[x_R^q;\hat{x}_H^q]\notin {X}_S\}$. 
Then the interactive constraint in (\ref{eq: benchmark smoothing chp 5}) is equivalent to $d^*(x_R^q,\mathcal{O}_q)\geq 0$ where $d^*(x_R^q,\mathcal{O}_q)$ is the signed distance function to $\mathcal{O}_q$ such that
\begin{eqnarray}
d^*(x_R^q,\mathcal{O}_q):=\left\{\begin{array}{cc}
\min_{z\in\partial\mathcal{O}_q}\|x_R^q-z\| & x_R^q\notin\mathcal{O}_q\\
-\min_{z\in\partial\mathcal{O}_q}\|x_R^q-z\| & x_R^q\in\mathcal{O}_q
\end{array}\right..
\end{eqnarray}
The symbol $\partial\mathcal{O}_q$ denotes the boundary of the obstacle $\mathcal{O}_q$. 

\begin{figure}[t]
\begin{center}
\subfloat[Cartesian space.\label{fig: cartesian}]{\makebox[3.5cm][c]{\includegraphics[width=3cm]{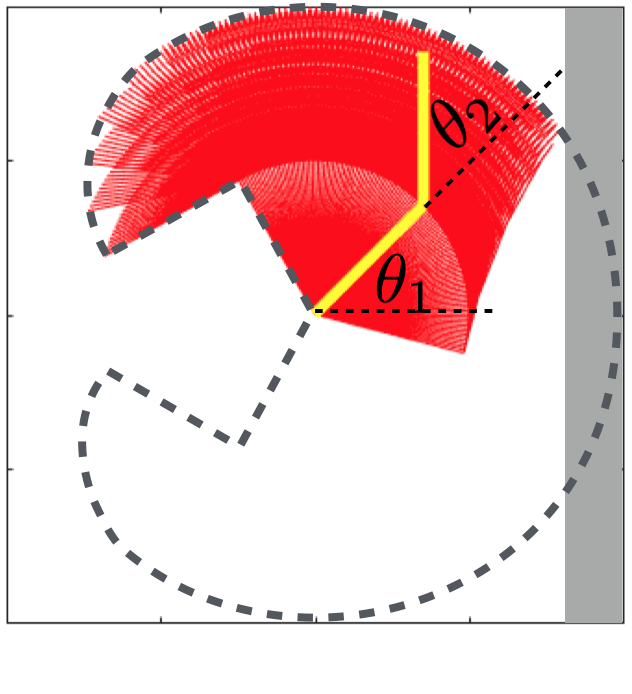}}}
\subfloat[Configuration space.\label{fig: configuration}]{\makebox[5cm][c]{\includegraphics[width=4.3cm]{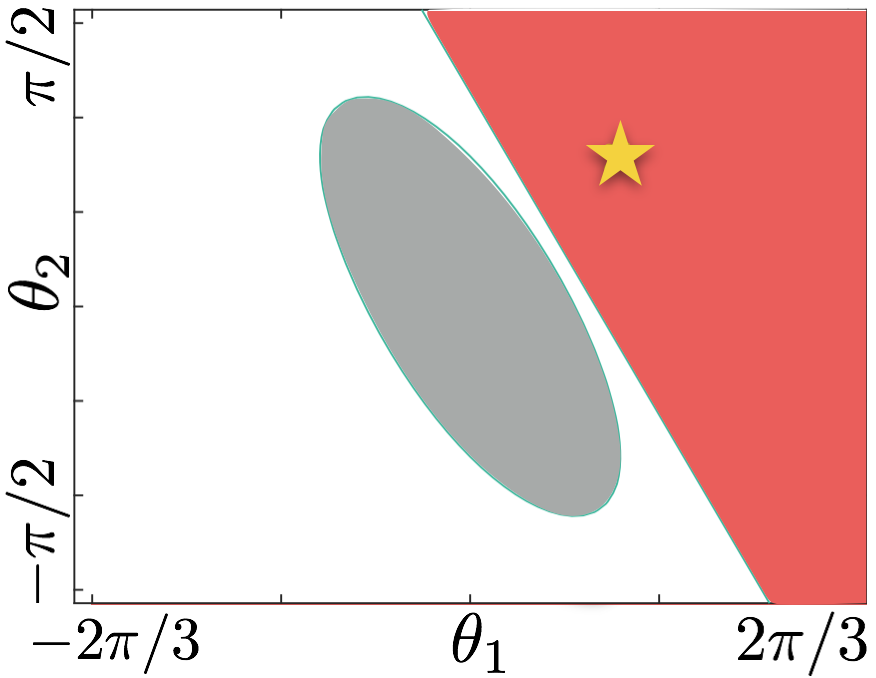}}}
\caption{Illustration of the constraint and the convex feasible set for a planar robot arm in Cartesian and Configuration spaces.}
\label{fig: c to c}
\end{center}
\end{figure}

If $\mathcal{O}_q$ is convex, then the function $d^*(\cdot,\mathcal{O}_q)$ is also convex. Hence $d^*(x_R^q,\mathcal{O}_q)\geq d^*(r^q,\mathcal{O}_q)+\nabla d^*(r^q,\mathcal{O}_q)(x_R^q-r^q)$ for any reference point $r^q$. Then $d^*(r^q,\mathcal{O}_q)+\nabla d^*(r^q,\mathcal{O}_q)(x_R^q-r^q)\geq 0$ implies that $x_R^q\notin \mathcal{O}_q$. If the obstacle $\mathcal{O}_q$ is not convex, we then break it into several simple overlapping convex objects $\mathcal{O}^i_q$ such as circles or spheres, polygons or polytopes. Then $d^*(\cdot,\mathcal{O}^i_q)$ is the convex cone of the convex set $\mathcal{O}_q^i$. Suppose the reference trajectory is $\mathbf{r}:=[r^0;r^1;\ldots;r^h]$, the convex feasible set $\mathcal{F}(\mathbf{r})$ for (\ref{eq: benchmark smoothing chp 5}) is defined as
\begin{subequations}
\begin{align}
& \mathcal{F}(\mathbf{r}):=\{\mathbf{x}_R: \mathcal{L}(\mathbf{x}_R)\in\oplus_h\Omega,x_R^q\in \Gamma,\\
&~~~~ d^*(r^q,\mathcal{O}_q^i)+\nabla d^*(r^q,\mathcal{O}_q^i)(x_R^q-r^q)\geq 0, \forall q,i\},\label{eq: convex distance}
\end{align}
\end{subequations}
which is a convex subset of $\Gamma^e$.

Starting from an initial reference trajectory $\mathbf{x}_R^{(0)}$, the  convex optimization (\ref{eq: cfs}) needs to be solved iteratively until either the solution converges or the descent of cost $J_R^d$ is small.
\begin{equation}
\mathbf{x}_R^{(k+1)} = \arg\min_{\mathbf{x}_R\in\mathcal{F}(\mathbf{x}_R^{(k)})} J_R^d(\mathbf{x}_R).\label{eq: cfs}
\end{equation}
It has been proved in \cite{liu2017sicon} that the sequence $\{\mathbf{x}_R^{(k)}\}$ converges to a local optimum of problem (\ref{eq: benchmark smoothing chp 5}). Moreover, the computation time can be greatly reduced using CFS. This is due to the fact that we directly search for solutions in the feasible area. Hence 1) the computation time per iteration is smaller than existing methods as no linear search is needed, and 2) the number of iterations is reduced as the step size (change of the trajectories between two consecutive steps) is unconstrained.

\subsection{Example: Long term planning for a planar robot arm}
The previous example in \Cref{fig: human robot cooperation} is considered. Let $T = ht_s$. Then the discretized optimization problem is formulated as 
\begin{subequations}\label{eq: discretize problem robot arm}
\begin{align}
\min_{\boldsymbol{\theta}} ~& \sum_q \|\theta^q - \mathcal{D}(p_l^q)\| + \|\frac{{\theta}^{q+1} - 2{\theta}^{q}+{\theta}^{q-1}}{t_s^2}\|,\label{eq: discretize cost}\\
s.t. ~& \theta_1^q\in[-2\pi/3,2\pi/3],\theta_2^q\in[-\pi/2,\pi/2],\label{eq: discretize state constraint}\\
&  \|p_r^q-p\|\geq d_{min},\forall p\in\mathcal{C}(\theta^q),\forall q,\label{eq: discretize safety constraint}
\end{align}
\end{subequations}
where $\boldsymbol{\theta} = [\theta^0;\ldots;\theta^h]$, (\ref{eq: discretize cost}) is the discrete version of (\ref{eq: cost planar arm}) and (\ref{eq: discretize safety constraint}) is the discrete version of (\ref{eq: safety constraint planar arm}). The control input $\ddot \theta$ is computed using finite difference. The constraint (\ref{eq: discretize safety constraint}) is in the Cartesian space $\mathbb{R}^2$, which needs to be transformed into the configuration space $[-\pi,\pi)\times [-\pi,\pi)$. For example, suppose that the base of the robot arm is at the origin; the links of the robot are of unit length; and the constraint (\ref{eq: discretize safety constraint}) requires that the robot arm is outside of the danger zone $x\geq 1.7$, i.e. $[0~1]p\leq 1.7$. The danger zone is shown in gray in \Cref{fig: cartesian}. Then the constraint in the configuration space is $\cos\theta_1+\cos(\theta_1+\theta_2)\leq 1.7$ where the danger zone is transformed into a convex ellipsoid-like object as shown in \Cref{fig: configuration}. Consider a reference point $(\pi/4, \pi/4)$, the convex feasible set in the configuration space is computed as
$ \cos(\pi/4)+\cos(\pi/2) -\sin(\pi/4)(\theta_1-\pi/4)-\sin(\pi/2)(\theta_1+\theta_2-\pi/2)=\sqrt{2}/{2} - \sqrt{2}/{2}(\theta_1-\pi/4)-(\theta_1+\theta_2-\pi/2)\leq 1.7$, which is illustrated in the red area in \Cref{fig: configuration}. 
Note that we did not use a distance function to convexify the obstacle, since the function $\cos\theta_1+\cos(\theta_1+\theta_2)$ is already convex in the domain (\ref{eq: discretize state constraint}). The area occupied by the robot given the configuration in the convex feasible set is illustrated in red in \Cref{fig: cartesian}. The dashed curve denotes the reachable area of the robot arm given the constraint (\ref{eq: discretize state constraint}). The constraint (\ref{eq: discretize safety constraint}) can be convexified for every step $q$. Then the discrete long term problem (\ref{eq: discretize problem robot arm}) can be solved efficiently by iteratively solving a sequence of quadratic programs.



\section{Evaluation Platforms\label{sec: platforms}}

\begin{figure*}[t]
\begin{center}
\subfloat[Virtual reality-based human-in-the-loop platform.\label{fig: platform 1}]{\includegraphics[width=8cm]{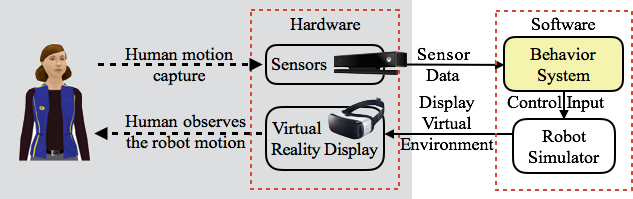}}~~
\subfloat[Dummy-robot interaction platform.\label{fig: platform 2}]{\includegraphics[width=9cm]{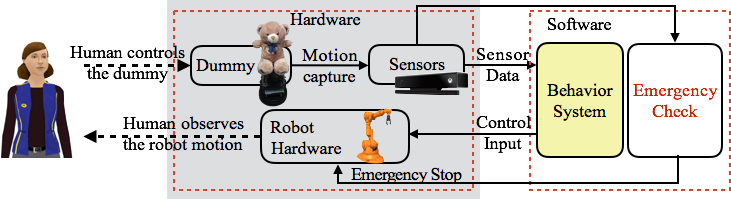}}\\
\caption{The evaluation platforms.}
\label{fig: platform}
\end{center}
\end{figure*}

To evaluate the performance of the designed behavior system, pure simulation is not enough. However, to protect human subjects, it is desired that we separate human subjects and robots physically during the early phase of deployment. This section introduces two evaluation platforms that we developed to evaluate the designed behavior system, as shown in \Cref{fig: platform}. 

\subsection{A Human-in-the-loop Simulation Platform\label{sec: hri kinect}}
The platform in \Cref{fig: platform 1} is a virtual reality-based human-in-the-loop platform. The robot's motion is simulated in the robot simulator. A human subject observes the robot movement through the virtual reality display (e.g. virtual reality glasses, augmented reality glasses or monitors). The reaction of the human subject is captured by sensors (e.g. Kinect or touchpad). The sensor data is sent to the behavior system to compute desired control input. The advantage of such platform is that it is safe to human subjects and convenient for idea testing. The disadvantage is that the robot simulator may neglect dynamic details of the physical robot, hence may not be reliable.
Nonetheless, the human data obtained in such platform can be stored for offline analysis in order to let the robot build better cognitive models.

\begin{figure}[t]
\begin{center}
\subfloat[Human-in-the-loop simulation platform.\label{fig: simulsys}]{
\includegraphics[width=5cm]{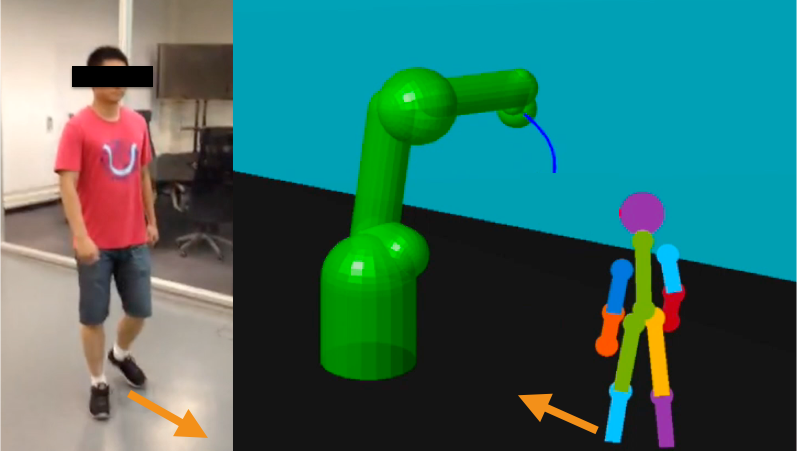}
}\\
\subfloat[Time step 20:5:40\label{fig: ssa 20-40}]{
\includegraphics[width=3cm]{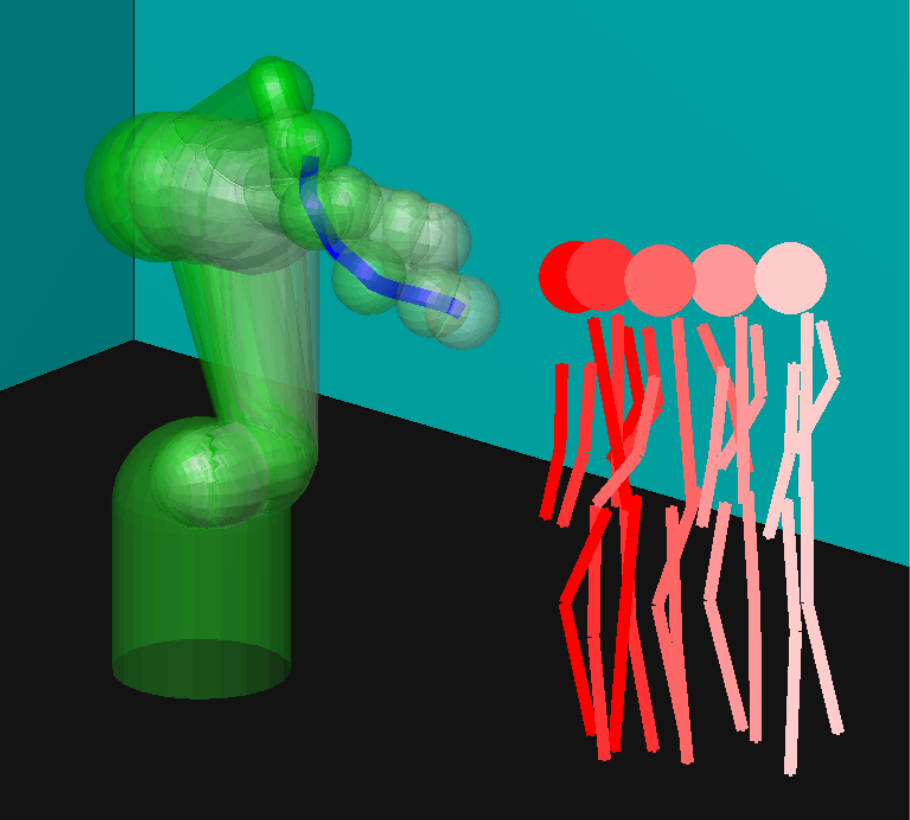}}
\subfloat[Time step 100:5:120\label{fig: ssa 100-120}]{
\includegraphics[width=3cm]{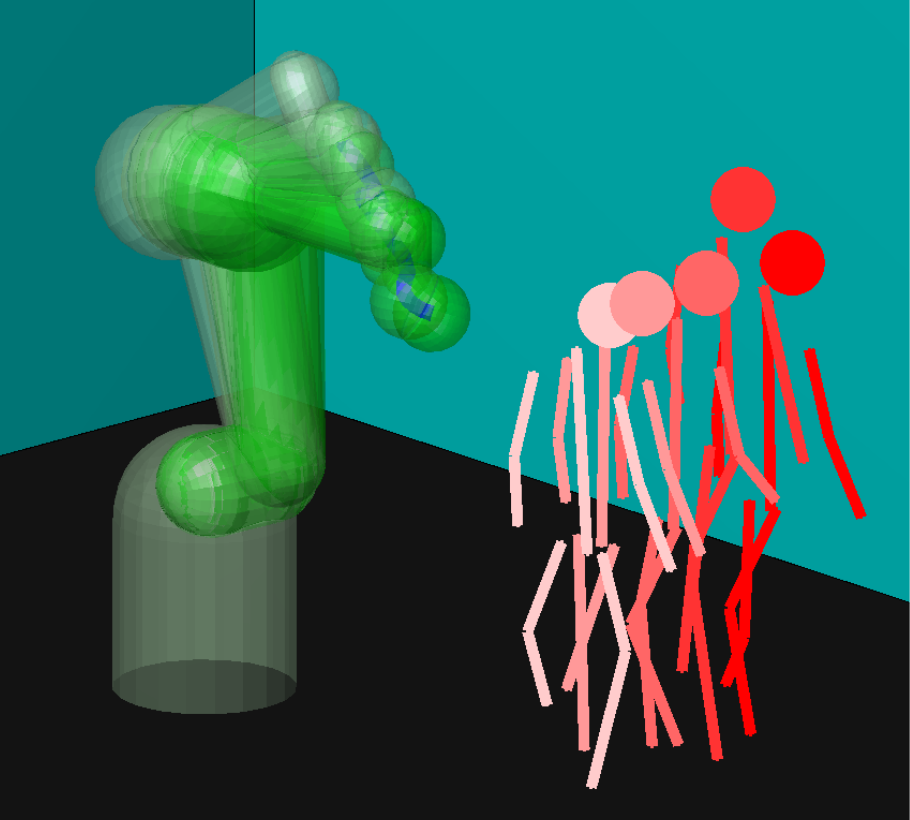}}
\caption{The human-in-the-loop simulation.}
\label{fig: ssa snapshot}
\vspace{-10pt}
\end{center}
\end{figure}

One human-in-the-loop simulation platform for industrial robot is illustrated in \Cref{fig: simulsys}. The left of the figure shows the human subject whose motion is captured by a Kinect. The right of the figure shows the virtual environment displayed on a screen. The robot links are replaces with capsules, while the human is displayed as skeletons. The task for the robot is relatively simple in this platform, e.g. to hold the neutral position or to follow a simple trajectory. 
The only hardware requirement is a computer, a Kinect and a monitor. The software package is available \texttt{github.com/changliuliu/VRsim}.

\subsection{A Dummy-Robot Interaction Platform\label{sec: dummy-robot}}
The platform in \Cref{fig: platform 2} is a dummy-robot interaction platform. The interaction happens physically between the dummy and the robot. The human directly observes the robot motion and controls the dummy to interact with the robot. As there are physical interactions, an emergency check module is added to ensure safety. The advantage of such platform is that it is safe to human subjects, while it is able to test interactions physically. However, the disadvantage is that the dummy usually doesn't have as many degrees of freedom as a human subject does.

One dummy-robot platform is shown in \Cref{fig: dummy-robot platform}, where the robot is performing a pick-and-place task. There is a 6 degree of freedom industrial robot arm (FANUC LR Mate 200iD/7L) with gripper, a Kinect for environment monitoring, a workpiece, a target box and an obstacle. The environment perceived by the robot is visualized on a screen. The status of the robot is shown in the monitor. The robot needs to pick the workpiece and place it in the target box while avoiding the dynamic obstacle. The scenario can be viewed as a variation of the human-robot cooperation in \Cref{fig: human robot cooperation}, where the green target box represents the human's left hand that accepts the workpiece from the robot, and the red box represents the human's right hand that needs to be avoided. The positions of the boxes can be controlled by human subjects at a distance to the robot arm.

\section{Experiment Results\label{sec: result}}

\subsection{The Human-in-the-loop Platform} 
\begin{figure}[t]
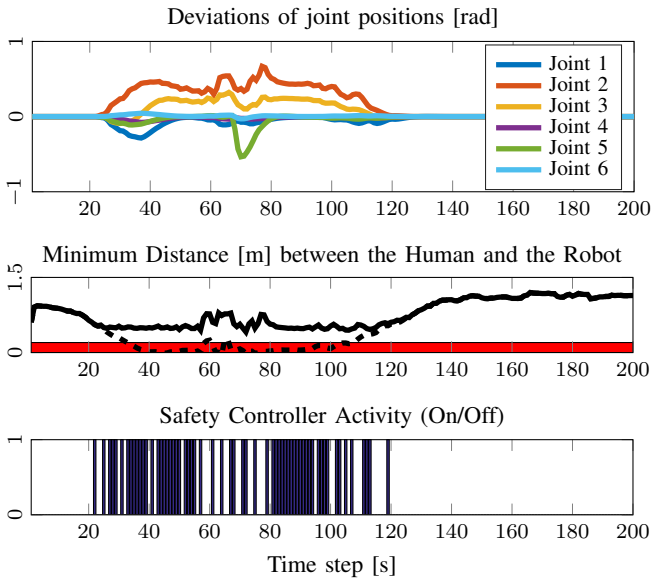

\begin{center}
\input{src/rsis-figure.tex}\\
\vspace{5pt}
\input{src/rsis-distance.tex}\\
\vspace{5pt}
\begin{tikzpicture}%
\begin{axis}[%
width=8cm,
height=1cm,
at={(1.011in,0.642in)},
scale only axis,
bar shift auto,
xmin=1,
xmax=200,
font = \footnotesize,
xlabel style={font=\small},
xlabel={Time step [\si{\second}]},
ymin=0,
ymax=1,
yticklabel style={rotate=90},
ytick={0,1},
axis background/.style={fill=white},
title style={yshift = -5, font=\small},
title={Safety Controller Activity (On/Off)}
]
\addplot[ybar, bar width=0.8, fill=mycolor7, area legend] table[row sep=crcr] {%
1	0\\
2	0\\
3	0\\
4	0\\
5	0\\
6	0\\
7	0\\
8	0\\
9	0\\
10	0\\
11	0\\
12	0\\
13	0\\
14	0\\
15	0\\
16	0\\
17	0\\
18	0\\
19	0\\
20	0\\
21	0\\
22	1\\
23	0\\
24	0\\
25	1\\
26	0\\
27	1\\
28	1\\
29	1\\
30	0\\
31	1\\
32	0\\
33	1\\
34	1\\
35	1\\
36	1\\
37	1\\
38	1\\
39	1\\
40	0\\
41	1\\
42	0\\
43	1\\
44	1\\
45	1\\
46	1\\
47	1\\
48	1\\
49	1\\
50	1\\
51	0\\
52	1\\
53	1\\
54	1\\
55	1\\
56	0\\
57	1\\
58	0\\
59	0\\
60	0\\
61	1\\
62	0\\
63	0\\
64	1\\
65	0\\
66	0\\
67	1\\
68	1\\
69	0\\
70	0\\
71	1\\
72	1\\
73	0\\
74	0\\
75	1\\
76	0\\
77	0\\
78	0\\
79	1\\
80	0\\
81	1\\
82	1\\
83	1\\
84	1\\
85	1\\
86	1\\
87	1\\
88	1\\
89	1\\
90	1\\
91	1\\
92	1\\
93	1\\
94	1\\
95	0\\
96	1\\
97	1\\
98	1\\
99	1\\
100	0\\
101	0\\
102	1\\
103	1\\
104	0\\
105	1\\
106	0\\
107	1\\
108	0\\
109	0\\
110	0\\
111	1\\
112	1\\
113	1\\
114	0\\
115	0\\
116	0\\
117	0\\
118	0\\
119	1\\
120	0\\
121	0\\
122	0\\
123	0\\
124	0\\
125	0\\
126	0\\
127	0\\
128	0\\
129	0\\
130	0\\
131	0\\
132	0\\
133	0\\
134	0\\
135	0\\
136	0\\
137	0\\
138	0\\
139	0\\
140	0\\
141	0\\
142	0\\
143	0\\
144	0\\
145	0\\
146	0\\
147	0\\
148	0\\
149	0\\
150	0\\
151	0\\
152	0\\
153	0\\
154	0\\
155	0\\
156	0\\
157	0\\
158	0\\
159	0\\
160	0\\
161	0\\
162	0\\
163	0\\
164	0\\
165	0\\
166	0\\
167	0\\
168	0\\
169	0\\
170	0\\
171	0\\
172	0\\
173	0\\
174	0\\
175	0\\
176	0\\
177	0\\
178	0\\
179	0\\
180	0\\
181	0\\
182	0\\
183	0\\
184	0\\
185	0\\
186	0\\
187	0\\
188	0\\
189	0\\
190	0\\
191	0\\
192	0\\
193	0\\
194	0\\
195	0\\
196	0\\
197	0\\
198	0\\
199	0\\
200	0\\
201	0\\
202	0\\
203	0\\
204	0\\
205	0\\
};
\addplot[forget plot, color=white!15!black] table[row sep=crcr] {%
1	0\\
200	0\\
};
\end{axis}
\end{tikzpicture}%
\caption{The result in the human-in-the-loop simulation}
\label{fig: ssa performance}
\end{center}
\vspace{-10pt}
\end{figure}

The experiment with the human-in-the-loop platform was run on a MacBook Pro with 2.3 GHz Intel Core i7. The robot arm is required to stay in the neutral position while maintaining a minimum distance $d_{min}=0.2m$ to the human. The sampling time is $t_s=0.1s$. The human state $x_H$ is taken as the skeleton position of the human subject. 
The uncertainty of human movement is bounded by a predefined maximum speed of the human subject. More accurate prediction models are discussed in \cite{liu2018serocs}.
The robot in the simulation has six degree of freedoms. Its control input is the joint acceleration $\ddot\theta$. Since the robot is not performing any task, a stabilizing controller that keeps the robot arm at the neutral position is adopted instead of a long term planner. Collision avoidance is taken care of by the short term safety controller, which implements the safe set algorithm discussed in \Cref{sec: ssa}. The safety index is the same as the one in \Cref{sec: ssa example}, which depends not only on the relative distance but also the relative velocity. 

The performance of the safety controller is shown in \Cref{fig: ssa performance}, where the first  graph shows the deviations of the joint positions from the neutral position, followed by a graph of the minimum distance between the human and the robot, and a graph of the activity of the safety controller. The horizontal axis is the time axis. In the distance profile, the solid curve is the true distance, while the dashed curve shows the distance in the case that the robot does not turn on the safety controller. Without the safety controller, the distance profile enters the danger zone ($d<d_{min}$) between time step 30 and 110. With the safety controller, the distance profile is always maintained above the danger zone. Moreover, as the dangerous situation is anticipated in advance, a modification signal is generated as early as at time step 22. The scenario at time step 20 to 40 is shown in \Cref{fig: ssa 20-40} where the snapshots are taken every 5 time steps. The lighter the color, the earlier it is in time. The blue curve is the trajectory of the robot end point. As the robot anticipates that the human will come closer, it moves backward. The scenario at time step 100 to 120 is shown in \Cref{fig: ssa 100-120}. As the robot anticipates that the human will go away, it returns to the neutral position. The safety controller is active between time step 20 to 120, which roughly aligns with the moments that the distance between the human and the robot in the neutral position is smaller than $d_{min}$. When the safety controller is active, all joint positions are affected.

The algorithm is run in Matlab script. The average computation time in solving (\ref{eq: convex optimization for local planning}) is 0.4ms, while the average time in obtaining $U_S^3$ is 3.9ms, where the computation of the distance $d$ takes the majority of the time. Nonetheless, the total computation time is still small enough for real time human robot interactions. Moreover, when compiled in C in the dummy-robot platform, the computation time can be kept lower than 1ms. 


\subsection{The Dummy-Robot Platform}

\begin{figure}[t]
\begin{center}
\subfloat[The platform.\label{fig: dummy-robot platform}]{
\includegraphics[width=7cm]{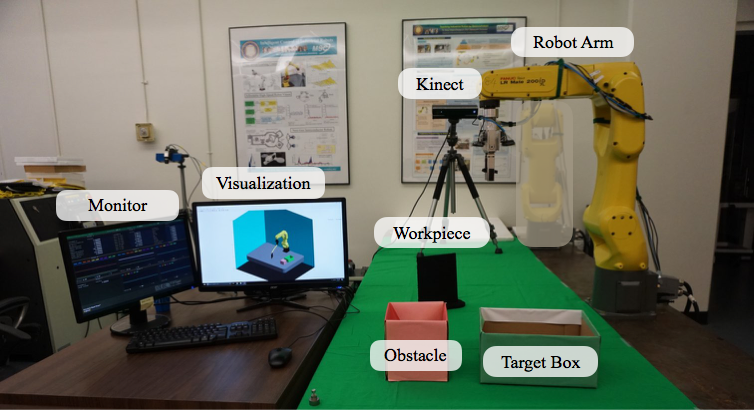}
}\\
\subfloat[RSIS software architecture.\label{fig: software architecture}]{\includegraphics[width=8.3cm]{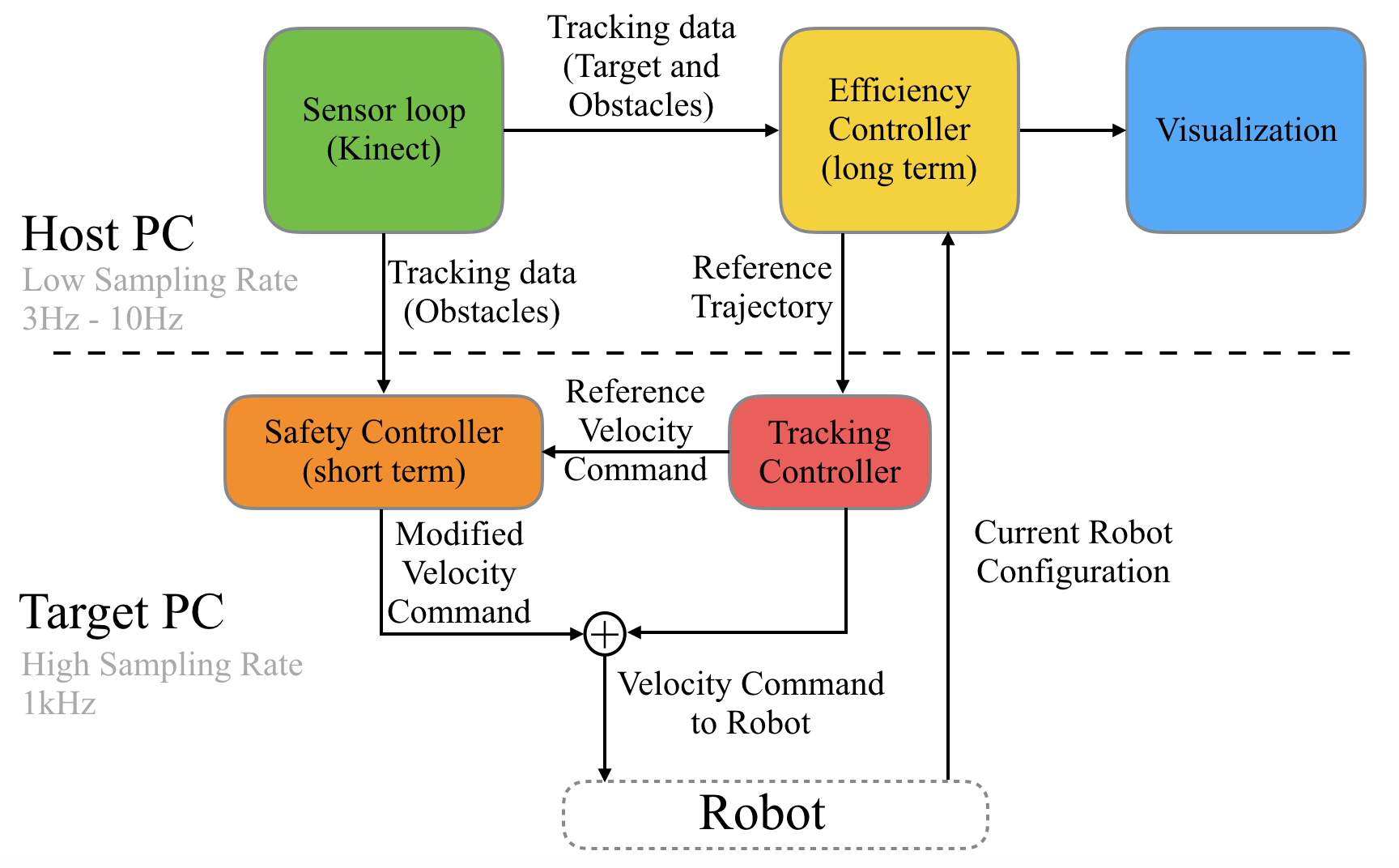}}
\caption{The dummy-robot platform for industrial robot.}
\end{center}
\end{figure}

\begin{figure*}
\begin{center}
\includegraphics[width=17cm]{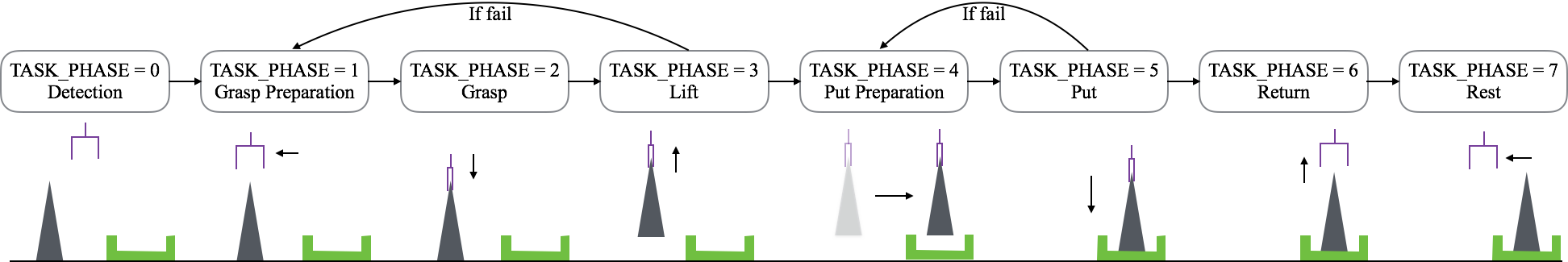}
\caption{Finite state machine for the pick-and-place task.}
\label{fig: finite state}
\end{center}
\end{figure*}

In the dummy-robot platform, the software implementation of the architecture in \Cref{fig: chap2 see think do} is shown in \Cref{fig: software architecture}. There are two PCs in the system. Algorithms that require longer computation time are implemented on the host PC, while algorithms that compute faster are implemented on the target PC. The target PC runs Simulink RealTime with Intel i5-3340 Quad-Core CPU \@ 3.10 GHz. Communications between the host PC and the target PC are enabled by user datagram protocol (UDP). 

There are five modules in the implementation: sensor loop, efficiency controller, visualization, safety controller and tracking controller. 

\subsubsection{Host PC} On host PC, sensor loop takes care of the perception and prediction in \Cref{fig: chap2 see think do}, which measures the state of the objects and predicts their future trajectories. In this platform, the trajectories of the objects are predicted assuming constant speed. The objects are attached with ARTags \cite{artag} for detection (now shown in the figure). The efficiency controller implements CFS to solve the following optimization problem,
\begin{subequations}\label{eq: problem dummy robot}
\begin{align}
\min_{\boldsymbol{\theta}}~~& J(\boldsymbol{\theta})=w_1 \|\boldsymbol{\theta}-\mathbf{r}\|^2_S+w_2 \|\boldsymbol{\theta}\|^2_R\\
s.t.~~& \theta^0 = r^0, \theta^h = r^h,\underline{\theta}\leq \theta^q/t_s\leq \bar\theta\label{eq: dummy robot constraint 1}\\
& \|p_r^q-p\|\geq d_{min}, \forall p\in\mathcal{C}(\theta^q),\forall q,\label{eq: dummy robot nonlinear constraint}
\end{align}
\end{subequations}
where $\boldsymbol{\theta}\in\mathbb{R}^{6(h+1)}$ is the discrete trajectory which consists of the joint positions of the robot in different time steps. The horizon $h$ is determined by the horizon of the reference trajectory $\mathbf{r}\in\mathbb{R}^{6(h+1)}$. 

The reference trajectory is generated according to the current task phase as specified in the finite state machine
in \Cref{fig: finite state}. Eight phases have been identified in the put-and-place task. 
In each phase, the ending pose will be estimated first. For example, by the end of the grasp preparation phase, the gripper should locate right above the workpiece. Then the time $T$ for the robot to go to the ending pose will be estimated assuming that the robot end point is moving at a constant speed. Then $h$ is set as $T/t_s$, $r^0$ in the reference trajectory as the current pose, and $r^h$ as the ending pose. The reference trajectory $\mathbf{r}$ is then obtained by linear interpolation between $r^0$ and $r^h$. In the constraint (\ref{eq: dummy robot constraint 1}), the boundary constraints are specified, and $\underline{\theta}$ and $\bar{\theta}$ are the lower and upper joint limits. The collision avoidance constraint is specified in (\ref{eq: dummy robot nonlinear constraint}) where $p_r^q$ is the position of the obstacle at time step $q$ and $d_{min}\in\mathbb{R}^+$. To simplify the distance function, the robot geometry $\mathcal{C}(\theta^q)$ is represented by several capsules as shown in \Cref{fig: simulsys}. When the robot is carrying the workpiece, the workpiece is also wrapped in the capsules. The minimum distance $d$ is then computed with respect to the centerlines of the capsules as discussed in \cite{liu2016algorithmic}. 

The collision avoidance constraint (\ref{eq: dummy robot nonlinear constraint}) can be convexified according to (\ref{eq: convex distance}) where the gradients are taken numerically. The optimization problem is then solved using CFS. The final trajectory is sent to the target PC as a reference trajectory. 

\subsubsection{Target PC}
On target PC, a tracking controller is used to track the trajectory in order to account for delay, package loss, and sampling time mismatch between the two PCs. The safety controller monitors the trajectory. The safety index is the same as the one discussed in the example in \Cref{sec: ssa}. The uncertainty cone is bounded by the maximum acceleration of the object. Finally, a velocity command will be sent to a low level controller to move the robot. The algorithms in the host PC are run in Matlab script on the host PC, while the algorithms on the target PC are compiled into C.

\subsubsection{Performance}
\begin{figure*}
\begin{center}
\subfloat[TASK\_PHASE = 1. $T = 17s$.\label{fig: result 1}]{\includegraphics[width=5.6cm]{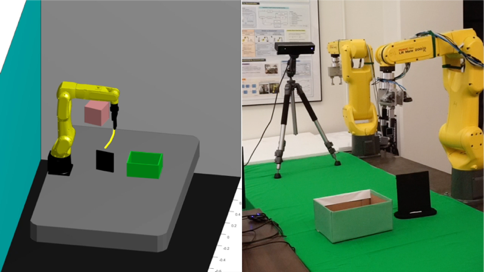}}~
\subfloat[TASK\_PHASE = 2. $T = 20s$.\label{fig: result 2}]{\includegraphics[width=5.6cm]{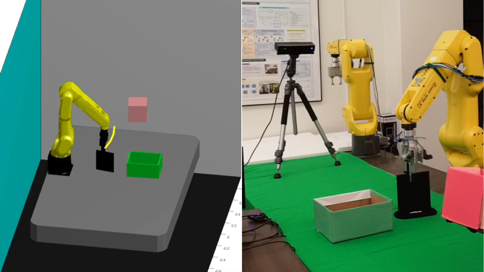}}~
\subfloat[TASK\_PHASE = 3. $T = 24s$.\label{fig: result 3}]{\includegraphics[width=5.6cm]{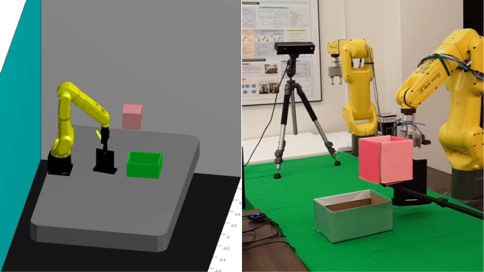}}\\
\subfloat[TASK\_PHASE = 3. $T = 26s$.\label{fig: result 4}]{\includegraphics[width=5.6cm]{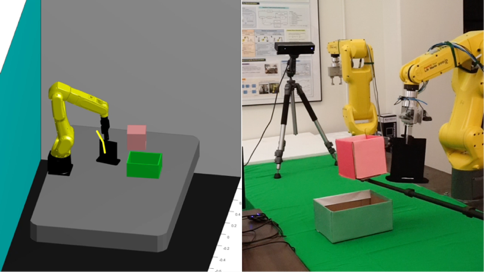}}~
\subfloat[TASK\_PHASE = 4. $T = 28s$.\label{fig: result 5}]{\includegraphics[width=5.6cm]{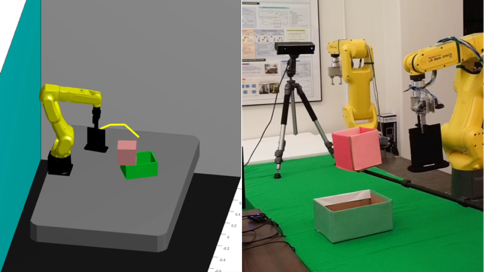}}~
\subfloat[TASK\_PHASE = 4. $T = 31s$.\label{fig: result 6}]{\includegraphics[width=5.6cm]{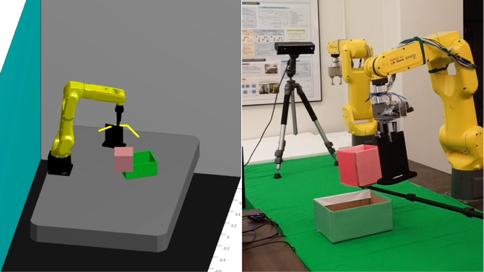}}\\
\subfloat[TASK\_PHASE = 5. $T = 34s$.\label{fig: result 7}]{\includegraphics[width=5.6cm]{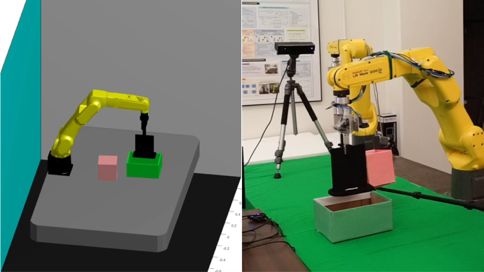}}~
\subfloat[TASK\_PHASE = 6. $T = 39s$.\label{fig: result 8}]{\includegraphics[width=5.6cm]{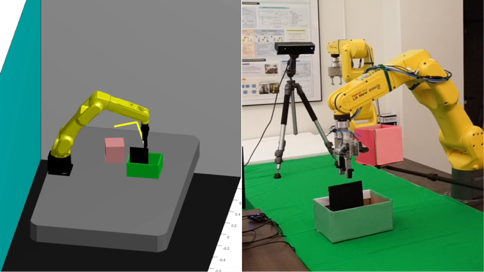}}~
\subfloat[TASK\_PHASE = 7. $T = 44s$.\label{fig: result 9}]{\includegraphics[width=5.6cm]{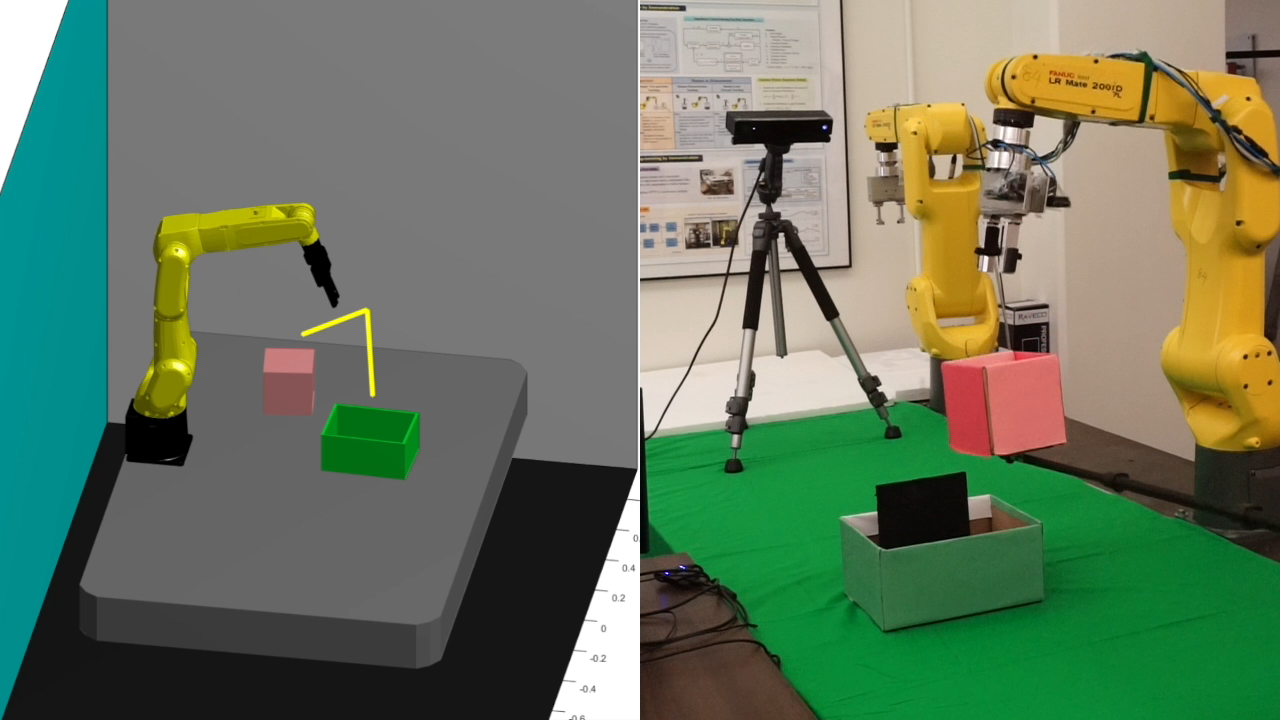}}
\caption{The experiment result on the dummy-robot platform.}
\label{fig: result}
\end{center}
\end{figure*}

The performance of the robot in one task scenario is shown in \Cref{fig: result}, where the left of every figure shows the perceived environment from the Kinect, while the right of every figure shows the real world scenario. The yellow curve is the planned long term trajectory. Though the trajectory is planned in the joint space, it is only shown for the robot end point for clarity. The obstacle is moving around causing disturbances to the robot arm. The robot generates detours to avoid the obstacle in TASK\_PHASE 4 as shown in \Cref{fig: result 5} and in TASK\_PHASE 6 as shown in \Cref{fig: result 8}. Sometimes the planned trajectory is perfectly followed as shown in \Cref{fig: result 6}. Sometimes the trajectory is modified by the safety controller to avoid the obstacle as shown in \Cref{fig: result 4,fig: result 9}, since the obstacle's state is not the same as predicted in the efficiency controller. 

To reduce the computation time on the host PC, the planning horizon is limited to $h\leq 6$. For solving the optimization \eqref{eq: problem dummy robot} in the efficiency controller, CFS takes 0.341s on average, while SQP takes 3.726s and  ITP takes 2.384s. CFS is significantly faster than SQP and ITP. In CFS algorithm, 98.1\% of the computation is used to convexify the problem and only 1.9\% of the time is used to solve the resulting quadratic programming. The most time consuming operation in convexification is to take the numerical derivative. Hence if better method is developed, the computation time can be further reduced. 


\section{Discussion\label{sec: discussion}}
It has been verified that RSIS is capable of interacting safely and efficiently with humans. Nonetheless, there are many directions for future works.

\subsubsection{To learn skills}
When the complexity of tasks increases, it is hard to design \textit{knowledge}, especially the internal cost \eqref{eq: chap2 design of knowledge} from scratch. It may be better for the robot to learn the skills directly from human demonstration \cite{tang2016teach}. 

\subsubsection{To consider communication among agents} This paper only optimizes physical movements of robots to accomplish tasks. To better facilitate interactions, intentions could also be communicated among agents to avoid mis-understandings. Moreover, physical movements can be communicative as it reveals one's intention to observers \cite{Dragan:2015aa}. We will build such knowledge about communicative motions into the robot behavior system in the future. 

\subsubsection{To improve computation efficiency} Computation efficiency of the designed algorithms is extremely important. This paper adopts convexification methods to speed up the computation of non-convex optimization. However, several assumptions narrowed the scope of the proposed algorithms, which should be relaxed in the future. 

\subsubsection{To analyze the designed robot behavior} We provided experimental verification of the designed robot behavior. However, theoretical analysis and formal guarantee is important to provide a deep insight into the system properties and ways to improve the design. In theoretical analysis, the questions that need to be answered are: (1) Is the parallel planner stable and optimal? (2) Will the learning process converge to the true value? (3) Will the designed behavior help achieve system optima in the multi-agent system? 

Based on RSIS, we recently developed safe and efficient robot collaboration system (SERoCS) \cite{liu2018serocs}, which further incorporates robust cognition algorithms for environment monitoring, and efficient task planning algorithms for reference generations. As a future work, we will also compare our methods with other methods \cite{ROB-052} on our platforms for a variety of tasks that involve human-robot interactions.

\section{Conclusion\label{sec: conclusion}}
This paper introduced the robot safe interaction system (RSIS) for safe real time human-robot interaction. First, a general framework concerning macroscopic multi-agent system and microscopic behavior system was discussed. The design of the behavior system was challenging due to uncertainties in the system and limited computation capacity. To equip the robot with a global perspective and ensure timely responses in emergencies, a parallel planning and control architecture was proposed, which consisted of an efficiency-oriented long term planner and a safety-oriented short term planner. The convex feasible set algorithm was adopted to speed up the computation in the long term motion planning. The safe set algorithm was adopted to guarantee safety in the short term. To protect human subjects in the early phase of deployment, two kinds of platforms to evaluate the designed behavior was proposed. The experiment results with the platforms verified the effectiveness and efficiency of the proposed method. Several future directions were pointed out in the discussion session and will be pursued in the future. 


\section*{Acknowledgement}
The authors would like to thank Dr. Wenjie Chen for the insightful discussions, and Hsien-Chung Lin and Te Tang for their helps with the experiment.

\end{document}